\pdfoutput=1

\documentclass[11pt]{article}

\usepackage[]{acl}

\usepackage{times}
\usepackage{latexsym}

\usepackage[T1]{fontenc}

\usepackage[utf8]{inputenc}

\usepackage{microtype}
\usepackage{graphicx}
\usepackage{multirow}
\usepackage{enumitem}
\usepackage{booktabs}
\usepackage{listings}
\usepackage{amsmath, bm}
\usepackage{pifont}

\makeatletter
    \newcommand\crulefill{\leavevmode
    \begingroup 
    \setlength{\dimen@}{0.5ex}
    \addtolength{\dimen@}{0.4pt}
    \leaders\hrule height\dimen@ depth -0.5ex \hfill
    \endgroup
    \kern\z@}

\title{JSEEGraph: Joint Structured Event Extraction as Graph Parsing}

\author{Huiling You$^1$, Samia Touileb$^2$ \and Lilja Øvrelid$^1$ \\
         $^1$University of Oslo\\
         $^2$University of Bergen \\ 
         \texttt{\{huiliny, liljao\}@ifi.uio.no} \\
         \texttt{samia.touileb@uib.no}
}

\begin{document}
\maketitle
\begin{abstract}
We propose a graph-based event extraction framework JSEEGraph that approaches the task of event extraction as general graph parsing in the tradition of Meaning Representation Parsing. It explicitly encodes entities and events in a single semantic graph, and further has the flexibility to encode a wider range of additional IE relations and jointly infer individual tasks. JSEEGraph performs in an end-to-end manner via general graph parsing: (1) instead of flat sequence labelling, nested structures between entities/triggers are efficiently encoded as separate nodes in the graph, allowing for nested and overlapping entities and triggers; (2) both entities, relations, and events can be encoded in the same graph, where entities and event triggers are represented as nodes and entity relations and event arguments are constructed via edges;  (3) joint inference avoids error propagation and enhances the interpolation of different IE tasks. We experiment on two benchmark datasets of varying structural complexities; ACE05 and Rich ERE, covering three languages: English, Chinese, and Spanish. Experimental results show that JSEEGraph can handle nested event structures, that it is beneficial to solve different IE tasks jointly, and that event argument extraction in particular benefits from entity extraction. Our code and models are released as open-source\footnote{\url{https://github.com/huiling-y/JSEEGraph}}.
\end{abstract}

\section{Introduction}
\begin{figure}[t!]
\centering
\includegraphics[width=\columnwidth]{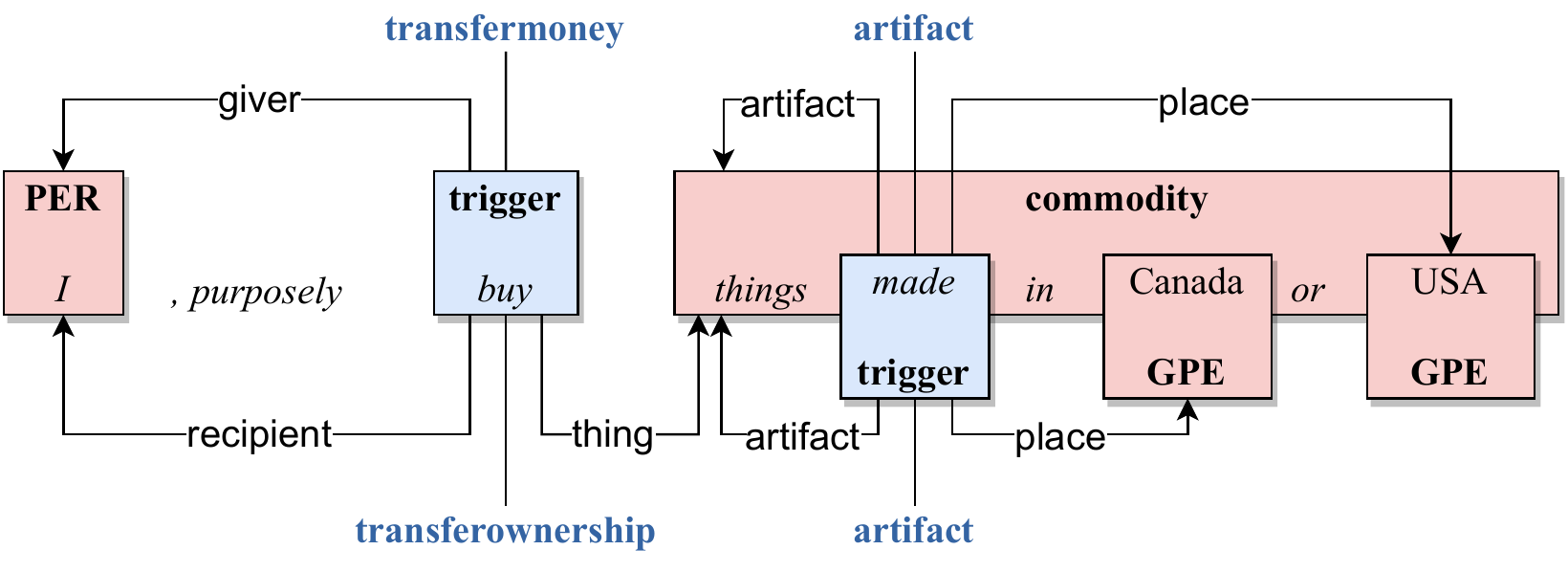}
\caption{\label{example} \footnotesize Example of nested and overlapping events in the sentence  \textit{``I, purposely buy things made in Canada or USA.'', taken from Rich ERE \citep{song-etal-2015-light}.}}
\end{figure}

Event extraction (EE) deals with the extraction of complex, structured representations of events from text, including overlapping and nested structures \cite{sheng-etal-2021-casee, cao-etal-2022-oneee}. While there are existing datasets annotated with such rich representations \citep{doddington-etal-2004-automatic, song-etal-2015-light}, a majority of current approaches model this task using simplified versions of these datasets or sequence-labeling-based encodings which are not capable of capturing the full complexity of the events.
Figure \ref{example} shows an example from the Rich ERE dataset \citep{song-etal-2015-light} of a sentence containing both nested and overlapping events: \textit{``buy''} serves as trigger for two overlapping events, \texttt{transfermoney} and \texttt{transferownership} with their respective argument roles, and similarly \textit{``made''} for two \texttt{artifact} events triggered by the coordination of two GPE entities \textit{Canada} and \textit{USA}; at the same time, the event trigger \textit{``made''} is nested inside the entity span \textit{``things made in Canada or USA''}. For this example, models based on token tagging (such as the commonly used BIO-encoding) would fail completely when a token contributes to multiple information extraction elements.
In this case, the version of the ACE05 dataset widely employed for EE would not fully capture the double-tagged event triggers, by simply disregarding one of the two events, and the nested entity \textit{``things made in Canada or USA''} would be \textit{``things''}.

Event extraction is a subtask of a wider set of Information Extraction (IE) tasks, jointly dealing with extracting various types of structured information from unstructured texts, from named entities, relations, to events. There have been continued efforts in creating benchmark datasets that can be used for evaluating a wide range of IE tasks. Both ACE05 \citep{doddington-etal-2004-automatic}\footnote{\url{https://catalog.ldc.upenn.edu/LDC2006T06}} and Rich ERE \citep{song-etal-2015-light}\footnote{\url{https://catalog.ldc.upenn.edu/LDC2020T18}} provide consistent annotations of entities, relations, and events. While there are clear inter-relations between these different elements, and despite the availability of rich annotations, existing works often deal with individual tasks, such as named entity recognition (NER) \citep{chiu-nichols-2016-named, bekoulis-etal-2018-adversarial} or event extraction (EE) \citep{yang-mitchell-2016-joint, du-cardie-2020-event, li-etal-2020-event}. Recently there have been some efforts in jointly modelling multiple IE tasks \citep{wadden-etal-2019-entity, lin-etal-2020-joint, nguyen-etal-2022-joint}, but these methods explicitly avoid nested instances. 

\begin{figure}[t!]
\centering
\includegraphics[width=\columnwidth]{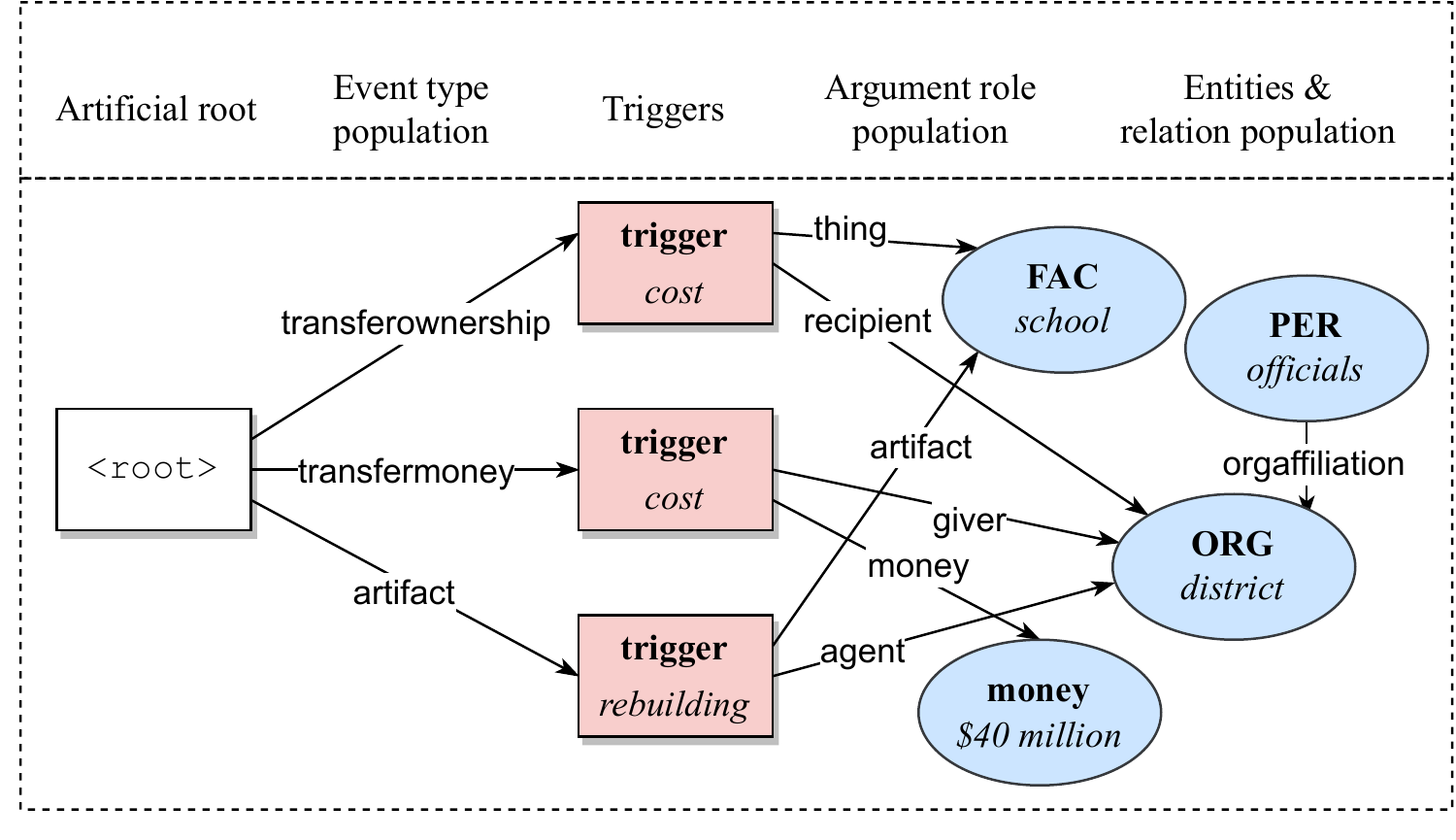}
\caption{\label{iegraph} \footnotesize Example of graph representation for entities, relations, and events from the sentence \textit{``School district officials have estimated the cost of rebuilding an intermediate school at \$40 million.''}, from Rich ERE~\citep{song-etal-2015-light}.}
\end{figure}

We here propose to represent events, along with entities and relations, as general graphs and approach the task of event extraction as Meaning Representation Parsing \citep{oepen-etal-2020-mrp, samuel-straka-2020-ufal}. 
As shown in Figure \ref{iegraph}, in such an information graph, event triggers and entities are represented as nodes; event types, argument roles, and relations are constrained edges; and nested/overlapped structures are straightforwardly represented, since a surface string can be abstracted into an unlimited number of nodes, as illustrated by the two separate nodes for the event triggers for \textit{``cost''}. Our approach does not rely on ontology- or language-specific features or any external syntactic/semantic parsers, but directly parses raw text into an information graph. 
We experiment on the benchmark datasets ACE05 \citep{doddington-etal-2004-automatic} and Rich ERE \citep{song-etal-2015-light}, zooming in on nested structures. Our results show JSEEGraph to be versatile in solving entity, relation, and event extraction jointly, even for heavily nested instances and across three different languages. Ablation studies consistently show that event extraction especially benefits from entity extraction.

The paper is structured as follows: section \ref{sec:related} provides the relevant background for our work, and section \ref{sec:task-data} further describes the tasks addressed and the datasets we employ, focusing in particular on their complexity, as measured by level of nesting. Section \ref{sec:parser} presents the JSEE graph parsing framework and section \ref{sec:exp} the experimental setup for evaluating the JSEE parser. Section \ref{sec:results} presents the results of our evaluations and provides a study of the performance for nested structures, as well as an ablation study assessing the effect of joint IE modeling and an error analysis. Finally we provide conclusions (Section \ref{sec:conclusion}) and discuss limitations of our work.

\section{Related work}
\label{sec:related}

Event extraction is commonly approached as supervised classification, even though other approaches relying \textit{e.g.} on generation \citep{paolini2021structured, lu-etal-2021-text2event, li-etal-2021-document, hsu2022degree} or prompt tuning inspired by natural language understanding tasks \citep{shin-etal-2020-autoprompt,gao-etal-2021-making,li-liang-2021-prefix,liu-etal-2022-dynamic} also are gaining ground. 
Classification-based methods break event extraction into several subtasks (trigger detection/classification, argument detection/classification), and either solve them separately in a pipeline-based manner \citep{ji-grishman-2008-refining, li-etal-2013-joint, liu-etal-2020-event, du-cardie-2020-event, li-etal-2020-event} or jointly infer them as multiple subtasks \citep{yang-mitchell-2016-joint, nguyen-etal-2016-joint-event, liu-etal-2018-jointly, wadden-etal-2019-entity, lin-etal-2020-joint}. Classification-based joint methods typically apply sequence-labeling-based encoding and extract all event components in one pass, whereas pipeline methods break the problem into separate stages which are performed sequentially. Whereas sequence-labeling approaches cannot distinguish overlapping events/arguments by the nature of the BIO-encoding, pipeline methods may in principle detect these. However, they typically suffer from error propagation and are not equipped to model the interactions between the different event elements (triggers, arguments).

\paragraph{Nested events} Some previous work addresses the problem of overlapping or nested arguments in EE.
\citet{xu-etal-2020-novel} address overlapping arguments in the Chinese part of the ACE05 dataset and jointly perform predictions for event triggers and arguments based on common feature representations derived from a pre-trained language model.
\citet{sheng-etal-2021-casee} propose a joint framework with cascaded decoding to tackle overlapping events, and sequentially perform type detection, event and argument extraction in a Chinese financial event dataset.
They deal with cases of both ``overlapping events'' and ``overlapping arguments'', however, their approach may suffer from error propagation due to the cascading approach.
\citet{cao-etal-2022-oneee} distinguish between overlapped and nested events and propose the OneEE tagging scheme which formulates EE as a word-to-word relation recognition, distinguishing separate span and role relations. OneEE is
evaluated on the FewFC Chinese financial event dataset and the biomedical event datasets Genia11 and Genia13. While specifically focusing on nested events, these previous works are limited by focusing only on one language or on specialized (financial/biomedical) domains. In this work we aim to provide a more comprehensive evaluation over two datasets in several versions with increasing levels of structural complexity (see below) and across three different languages.

\paragraph{Joint IE approaches}
\citet{wadden-etal-2019-entity} propose the DyGIE++ model which approaches joint modeling of IE entities and relations via span-based prediction of entities and event triggers, and subsequent dynamic graph propagation based on relations. They evaluate on ACE05 and Genia datasets and limit their experiments to English only. Their approach is restricted to a certain span width, limiting the length of possible entities.
OneIE \cite{lin-etal-2020-joint} is a joint system for IE using global features to model cross-subtask or cross-instance interactions between the subtasks and predict an information graph. They propose the E+ extension of ACE05 which includes multi-token events (E$^+$) as we do. As in our work, they also present results on Spanish and Chinese as well and develop a multilingual model, but their experiments avoid nested structures, by using only the head of entity mentions and specifically removing overlapped entities.
\citet{nguyen-etal-2022-joint} model joint IE in a two-stage procedure which first identifies entities and event triggers and subsequently classify relations between these starting from a fully connected dependency graph;
a GCN is employed to encode the resulting dependency graphs for computation of the joint distribution.
While the approach is shown to be effective, it is still a pipeline approach which can suffer from error propagation. Since it relies on 
sequence labeling for entity/event detection, it cannot identify overlapping entities/event triggers. Furthermore, the approach relies on syntactic information from an external parser and focuses only on English and Spanish in the Light ERE dataset \citep{song-etal-2015-light}.

\paragraph{Meaning Representation Parsing}
Meaning Representation Parsing (MRP)  \citep{oepen2014semeval,oepen2015semeval,oepen-etal-2020-mrp} is a framework covering several types of dependency-based semantic graph frameworks. Unlike syntactic dependency representations, these semantic representations are not trees, but rather general graphs, characterised by potentially having multiple top nodes (\textit{roots}) and not necessarily being connected, since not every token is necessarily a node in the graph. 
The semantic frameworks include representations with varying levels of ``anchoring'' to the input string \citep{oepen-etal-2020-mrp}, ranging from the so-called ``bi-lexical'' representations  where every node in the graph corresponds to a token in the input string to a framework like AMR \cite{banarescu2013abstract} which constitutes the most abstract and unanchored type of framework, such that the correspondence between the nodes in a graph and tokens in the string is completely flexible.
This allows for straightforward representation of nesting and overlapping structures, where multiple nodes may be anchored to overlapping sub-strings.
There have been considerable progress in developing variants of both transition-based and graph-based dependency parsers capable of producing such semantic graphs \cite{hershcovich-etal-2017-transition, dozat-manning-2018-simpler,samuel-straka-2020-ufal}. Previous research has further made use of AMR-based input representations to constrain the tasks of event extraction \cite{huang-etal-2018-zero} and more recently joint information extraction \cite{zhang-ji-2021-abstract}, where an off-the-shelf AMR parser is used to derive candidate enitity and event trigger nodes before classifying pairwise relations guided by the AMR hierarchical structure.
While there are clear parallels between the MRP semantic frameworks and the tasks proposed in IE, little work has focused on the direct application of MRP parsing techniques to these tasks.
\citet{you2022eventgraph} is a notable exception in this respect, who presents an adaptation of the PERIN semantic parser \cite{samuel-straka-2020-ufal} to the event extraction task. While their work is promising it is limited to only one dataset (ACE05), which does not contain a lot of nested structures and is further limited to English event extraction only. In this work we extend their approach to the task of joint information extraction, covering both entities, events and relations taken from two different datasets in several versions and for three languages, and further demonstrates the effectiveness of approaching general information extraction from text via graph-parsing and the interpolation of different IE tasks.

\section{Task and Data}
\label{sec:task-data}

While the main focus of this work is on event extraction, we hypothesize that our graph-based approach lends itself to dealing with two challenging aspects of current research on this task: the processing of nested and overlapping event structures, and the joint modeling of inter-related IE structures. In the following we quantify the level of nesting in two widely used datasets which contain rich annotations for both entities, events, and relations. We further propose two versions of each dataset with varying potential for nesting, which allows us to focus on this aspect during evaluation.

\paragraph{Event Extraction} is the task of extracting events into structured forms, namely event triggers and their arguments. An event trigger is the word(s) that most clearly describes an event, such as \textit{``buy''}, which evokes a \texttt{transferownership} and an \texttt{transfermoney} event in Figure \ref{example}. Event arguments are the participants and attributes of an event, and can be tagged as entities at the same time, as demonstrated in Figure \ref{iegraph}.

{\footnotesize
\begin{table}[t!]
\footnotesize
\resizebox{\columnwidth}{!}{%
\begin{tabular}{@{}ll@{\hspace{2em}}rrrrr@{} }
\toprule 
\textbf{Lang} & \textbf{Split} & \textbf{\#Sents} & \textbf{\#Events} & \textbf{\#Roles} & \textbf{\#Entities} & \textbf{\#Relations}\\
\midrule

\multicolumn{7}{@{}c@{}}{\raisebox{0.8ex}{\small\textbf{Dataset: ACE05 }  } } \\

\multirow{3}{*}{en} & Train & 19\,371 & 4\,419 & 6\,609 & 47\,546 & 7\,172 \\
& Dev & 896 & 468 & 759 & 3\,421 & 729 \\
& Test & 777 & 461 & 735 & 3\,828 & 822 \\ \midrule
\multirow{3}{*}{zh} & Train & 6\,706 & 2\,928 & 5\,576 & 29\,674 & 8\,003 \\
& Dev & 511 & 217 & 406 & 2\,246 & 601 \\
& Test & 521 & 190 & 336 & 2\,389 & 686 \\ \midrule

\multicolumn{7}{@{}c@{}}{\raisebox{0.8ex}{\small\textbf{Dataset: Rich ERE }  } } \\

\multirow{3}{*}{en} & Train & 12\,421 & 8\,368 & 15\,197 & 34\,611 & 7\,498 \\
& Dev & 692 & 459 & 797 & 1\,998 & 366 \\
& Test & 745 & 566 & 1\,195 & 2\,286 & 544 \\ \midrule
\multirow{3}{*}{zh} & Train & 9\,253 & 5\,325 & 9\,066 & 26\,128 & 6\,044 \\
& Dev & 541 & 366 & 522 & 1\,609 & 379 \\
& Test & 483 & 439 & 776 & 2\,022 & 502 \\ \midrule
\multirow{3}{*}{es} & Train & 8\,292 & 5\,013 & 8\,575 & 20\,347 & 4\,140 \\
& Dev & 383 & 254 & 447 & 1\,068 & 199 \\
& Test & 598 & 334 & 609 & 1\,438 & 287 \\
\bottomrule

\end{tabular}%
}
\caption{\label{tab:data-stats} \footnotesize Statistics of the preprocessed datasets.}
\end{table}}

\begin{table}[t]
\footnotesize
\resizebox{\columnwidth}{!}{%
\begin{tabular}{@{}l@{\hspace{2em}}rrrr@{}}
\toprule
\textbf{Dataset} & \textbf{\#Event-types} & \textbf{\#Argument-roles} & \textbf{\#Entity types} & \textbf{\#Relation type} \\ \midrule                  
ACE05 & 33 & 22 & 7 & 6 \\  
Rich ERE & 38 & 20 & 15 & 6 \\ \bottomrule
\end{tabular}%
}
\caption{\footnotesize Inventory of event types, argument roles, entity types and relation types in ACE05 and Rich ERE.}
\label{tab:ontology}
\end{table}

We use the benchmark datasets ACE05 \cite{doddington-etal-2004-automatic} and Rich ERE \cite{song-etal-2015-light}, both containing consistent annotations for entities, relations, and events, for joint evaluation of multiple IE tasks and in multiple languages (ACE05 in English and Chinese, and ERE in English, Chinese, and Spanish). Table \ref{tab:data-stats} summarizes the relevant statistics of the  datasets. The inventory of event types, argument roles, entity types and relation types are listed in Table \ref{tab:ontology}. Despite targeting the same IE tasks, from ACE05 to Rich ERE, the annotation guidelines have shifted towards more sophisticated representations, resulting in more complex structures in Rich ERE \cite{song-etal-2015-light}. Prominent differences between ACE05 and Rich ERE are:

\begin{itemize}[noitemsep]
    \item \textbf{Entities}, and hence event arguments, are more fine-grained in Rich ERE, with 15 entity types, as compared to 7 types in ACE05. In terms of entity spans, ACE05 explicitly marks the head of the entity versus the entire mention, providing the possibility of solving a simpler task for entity extraction and recognizing only the head token as opposed to the full span of the entity in question. This is commonly done for this task in previous work of EE. However, in Rich ERE, the entire string of text is annotated for entity mentions, and heads are only marked explicitly for nominal mentions that are not named entities or pronominal entities. 
    \item \textbf{Event triggers} can be double-tagged in Rich ERE, namely one trigger can serve multiple event mentions, giving rise to overlapping events, as shown in Figure \ref{example}, while in ACE05, an event trigger only evokes one event. This means that Rich ERE presents a more complex task of event extraction.
\end{itemize}

{\footnotesize
\begin{table}[t!]

\resizebox{\columnwidth}{!}{%
\begin{tabular}{@{}ll@{\hspace{2em}}ccc@{\hspace{2em}}cc@{} }
\toprule 
\multirow{2}{*}{\textbf{Dataset}} & \multirow{2}{*}{\textbf{Lang}} & \multicolumn{3}{@{}c@{\hspace{2em}}}{\textbf{Nesting}} & \multicolumn{2}{@{}c@{}}{\textbf{\#Sents}} \\

& & Trg-Trg & Ent-Ent & Trg-Ent & Nested & All \\ \midrule

\multirow{2}{*}{ACE05-E${^+}$} & en & 0 & 0 & 4 & 4 & 21044 \\

& zh & 0 & 4 & 9 & 12 & 7738\\ \midrule

\multirow{2}{*}{ACE05-E${^{++}}$} & en & 0 & 13387 & 716 & 5315 & 21044 \\

& zh & 0 & 10797 & 252 & 3748 & 7738 \\ \midrule

\multirow{3}{*}{Rich ERE-E${^+}$} & en & 1066 & 1329 & 244 & 1529 & 13858 \\

& zh & 301 & 1383 & 284 & 1266 & 10277 \\ 
& es & 485 & 523 & 97 & 712 & 9273\\ \midrule

\multirow{3}{*}{Rich ERE-E${^{++}}$} & en & 1063 & 9453 & 1517 & 4277 & 13858 \\

& zh & 301 & 7303 & 622 & 2993 & 10277 \\ 
& es & 485 & 5526 & 854 & 2614 & 9273 \\ \bottomrule

\end{tabular}%
}
\caption{\label{tab:nesting}  \footnotesize Nesting instances in ACE05 and Rich ERE. Nesting between a pair of event triggers is referred to as \texttt{Trg-Trg}; between a pair of entity mentions as \texttt{Ent-Ent}, and between an event trigger and an entity as \texttt{Trg-Ent}. For both datasets, in the E${^+}$ version, entity mentions include only heads, while in the E${^{++}}$ version, entity mentions include the full text spans.} 
\end{table}}

We measure the nested instances in ACE05 and Rich ERE as a way to showcase different levels of complexity for extracting entities, relations, and events. More specifically, we quantify nested instances in two versions of each dataset, one using only the head of an entity mention (when it is annotated), and the other with the entire mention text. Following \citet{lin-etal-2020-joint} we dub the version which only marks the head of entities ACE-E${^+}$ and Rich ERE-E${^+}$, and introduce two additional versions of the datasets, dubbed, ACE-E${^{++}}$ and Rich ERE-E${^{++}}$ which retain the full annotated mention text span. Nesting is measured between any pair of triggers and entities. Note that our notion of nesting subsumes both \textit{overlapping} and \textit{nested} target/entities \citep{cao-etal-2022-oneee}, \textit{i.e.} both full and partial overlap of text spans.  As shown in Table \ref{tab:nesting}, Rich ERE features many cases of nested triggers, while these are not found in ACE05, due to the aforementioned double-tagging in Rich ERE (see Figure \ref{example}); when only considering the head of an entity, ACE05 exhibits very little nesting, but Rich ERE exhibits a considerable amount of nesting within entities, as well as between entity-trigger. The reason for this is that in Rich ERE, only certain nominal mentions are marked with explicit heads; when the full entity mentions are considered, both datasets are heavily nested.

As mentioned above, this work deals with three IE tasks, as exemplified by Figure \ref{iegraph}: entities, relations, and events. Given a sentence, our JSEEGraph framework extracts its entity mentions, relations, and event mentions. In addition to event extraction, we thus target two additional IE tasks in our graph-based model:

\paragraph{Entity Extraction} is to identify entity mentions from text and classify them into types according to a pre-defined ontology. For example, in Figure \ref{iegraph}, \textit{``district''} is an organization (\texttt{ORG}) entity.

\paragraph{Relation Extraction} aims to assign a relation type to an ordered pair of entity mentions, based on a pre-defined relation ontology. For example, in Figure \ref{iegraph}, the relation between \texttt{PER} \textit{``officials''} and \texttt{ORG} \textit{``district''} is \texttt{orgaffiliation}.


\section{Graph parsing framework}
\label{sec:parser}

\begin{figure*}[t!]
\centering
\includegraphics[width=0.9\textwidth]{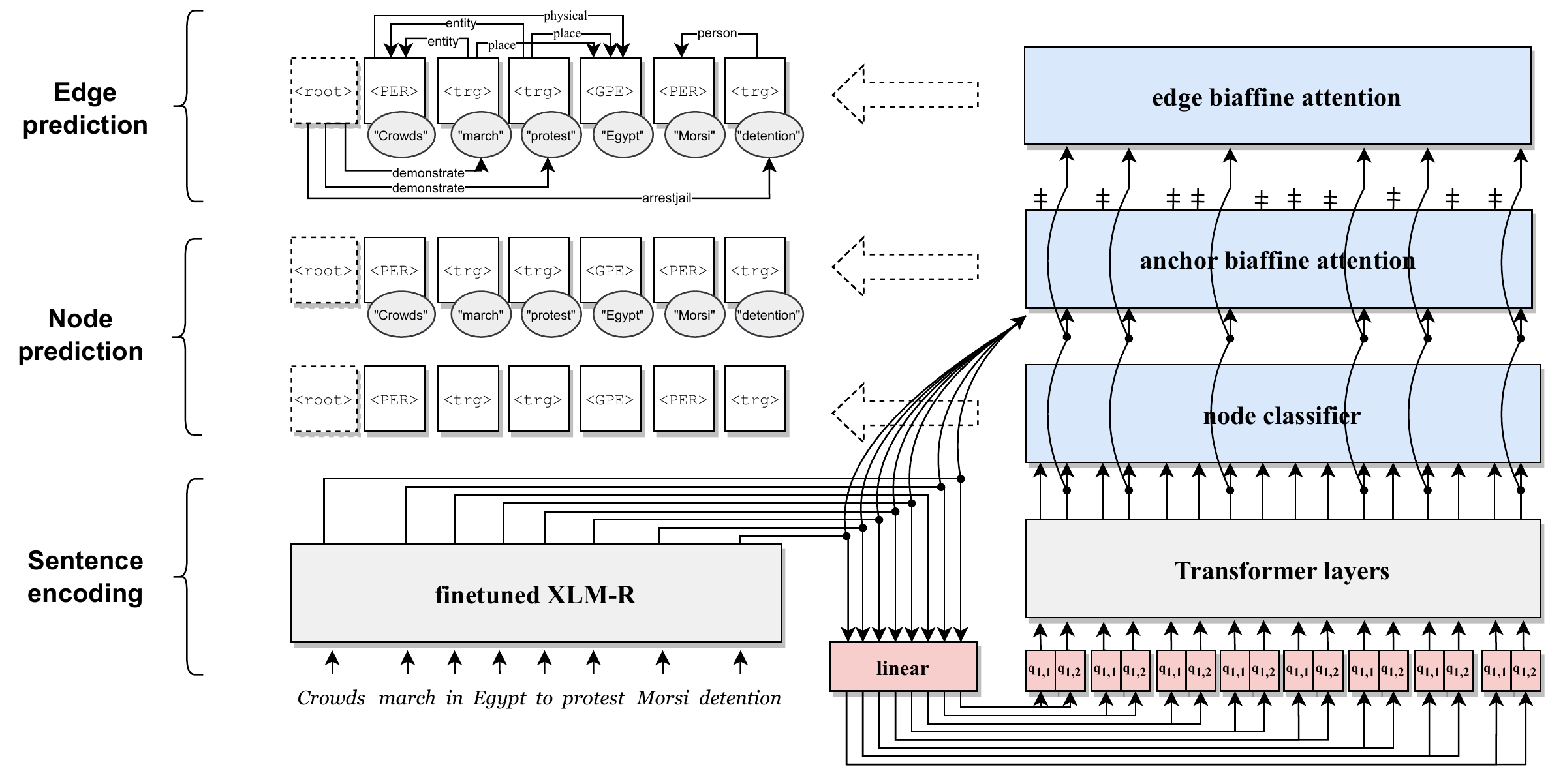}
\caption{\label{parser} \footnotesize An illustration of our JSEEGraph parsing the sentence \textit{``Crowds march in Egypt to protest Morsi detention.''}, example from Rich ERE. }
\end{figure*}

Our JSEEGraph framework is a text-to-graph parser tailored for EE tasks, additionally with different IE components explicitly encoded in a single graph, as shown in Figure \ref{iegraph}. Our framework builds on \citet{samuel-straka-2020-ufal} who developed the PERIN parser in the context of Meaning Representation Parsing \cite{oepen-etal-2020-mrp}, as well as \cite{you2022eventgraph} who applied PERIN to the task of event extraction. We here extend this parser to the IE graphs shown in Figure \ref{iegraph} in a multilingual setting. 

Given a sentence, as the example shown in Figure \ref{parser}, JSEEGraph encodes the input tokens with the pre-trained language model XLM-R \citep{conneau-etal-2020-unsupervised} to obtain the contextualized embeddings and further maps the embeddings onto queries; nodes (triggers and entities) are predicted by classifying the queries and anchored to surface tokens via a deep biaffine classifier \citep{dozat2017deep}; edges are constructed between nodes with two biaffine classifiers, assigning arguments to predicted events and relations to entity pairs. We describe each module in detail in what follows.

\subsection{Sentence encoding}

We use XLM-R \citep{conneau-etal-2020-unsupervised} to obtain the contextualized embeddings of the input sequence. To be specific, a trainable weight $w_l$ is used to get a weighted sum of representations of different layers, so the final contextual embedding $\mathrm{\textbf{e}} = \sum_{l=1}^L \mathrm{softmax}(w_l)\mathrm{\textbf{e}}_l$ with \textbf{e}$_l$ as the intermediate output from the $l^{th}$ layer. If an input token consists of multiple subwords, the final contextual embedding will be the weighted sum over all subword embeddings with a learned subword attention.

Each contextual embedding is mapped into \textbf{q} $=\{\mathrm{\textbf{q}}_1, \cdots, \mathrm{\textbf{q}}_n\}$ queries via a linear layer, and further transformed into hidden features \textbf{h} $=\{\mathrm{\textbf{h}}_1, \cdots, \mathrm{\textbf{h}}_n\}$ with a stack of transformer encoder layers, which models inter-query dependency with multi-head self-attention. 

\subsection{Node prediction}

The node prediction module consists of a node label classifier and an anchor biaffine attention classifier. 

The node label classifier is a linear classifier classifying each query into a node in the graph, and the node label is predicted by a single-layer feedforward network (FNN). If a query is classified into ``null'', no node is created from this query.

Node anchoring, as shown in Equation (\ref{eq:bilin}), is performed by biaffine attention \citep{dozat-manning-2018-simpler} between the contextual embeddings \textbf{e} and hidden feature of queries \textbf{h}, to map each query (a candidate node) to surface tokens, as shown in Equation (\ref{eq:node_anchor}). For each query, every input token is binary classified into anchor or non-anchor.

{\footnotesize
\begin{gather}
    \label{eq:bilin}
    \mathrm{Bilinear}(\mathrm{X}_1,\mathrm{X}_2) = \mathrm{X}_1^T\mathrm{UX}_2 \\
    \label{eg:biaffin}
    \mathrm{Biaffine}(\mathrm{X}_1,\mathrm{X}_2) = \mathrm{X}_1^T\mathrm{UX}_2 + \mathrm{W}(\mathrm{X}_1 \oplus \mathrm{X}_2) + b \\
    \label{eq:node_anchor}
    \mathrm{node}^{\mathrm{(anchor)}} = \mathrm{Biaffine}^{\mathrm{(anchor)}} (\mathrm{\textbf{h}}, \mathrm{\textbf{e}})
\end{gather}}

Node prediction is complete with queries that are classified into nodes and anchored  to corresponding surface tokens. Predicted nodes are either event triggers or entities, labeled as ``trigger'' or entity type. A dummy node is randomly generated to add to predicted nodes to play the role of \texttt{<root>} node, and always holds the first position. 

\subsection{Edge prediction}

Edge prediction between nodes is performed with two deep biaffine classifiers, as in Equation (\ref{eq:edge-cls}), one to predict edge presence between a pair of nodes and the other to predict the corresponding edge label. To construct edges between nodes, only queries from which nodes have been constructed will be used, and the new hidden features is \textbf{h}$^\prime$, which are further split into two parts with a single-layer FNN, as show in Equation (\ref{eq:edge-fnn1}) and (\ref{eq:edge-fnn2}).

{\footnotesize
\begin{gather}
    \label{eq:edge-fnn1}
    \mathrm{\textbf{h}}^{\prime (\mathrm{edge)}}_1 = \mathrm{FNN}^{(\mathrm{edge})}_1(\mathrm{\textbf{h}}^{\prime}) \\
    \label{eq:edge-fnn2}
    \mathrm{\textbf{h}}^{\prime (\mathrm{edge)}}_2 = \mathrm{FNN}^{(\mathrm{edge})}_2(\mathrm{\textbf{h}}^{\prime}) \\
    \label{eq:edge-cls}
    \mathrm{edge} =  \mathrm{Biaffine}^{\mathrm{(edge)}} (\mathrm{\textbf{h}}^{\prime (\mathrm{edge})}_1, \mathrm{\textbf{h}}^{\prime (\mathrm{edge})}_2) 
\end{gather}
}

 The edge presence biaffine classifier performs binary classification, deciding whether or not an edge should be constructed between a pair of nodes. The edge label biaffine classifier performs multi-class classification, and the edge label set is the union of argument roles and relation types.

 \subsection{Constrained decoding}
 
 During inference, we apply a set of  constraints specifically developed for the  correct treatment of event arguments and entity relations based on the graph encoding we define for the information graph (Figure \ref{iegraph}): 1) directed edges from the \texttt{<root>} node can only connect to a trigger node, and the corresponding edge label is an event type; 2) directed edges from a trigger node to an entity indicates an event argument, with the argument role placed as edge label; 3) directed edges between a pair of entities indicate an entity relation, and the corresponding relation type is assigned to the edge label. 

\section{Experimental setup}
\label{sec:exp}

\subsection{Data}
As mentioned above, we evaluate our system on the benchmark datasets   ACE05\footnote{\url{https://catalog.ldc.upenn.edu/LDC2006T06}} (LDC2006T06) and Rich ERE\footnote{\url{https://catalog.ldc.upenn.edu/LDC2020T18}} (LDC2020T18). As mentioned above, Table \ref{tab:data-stats} summarizes the statistics of the preprocessed datasets.

Following \citet{lin-etal-2020-joint}, we keep 33 event types, 22 argument roles, 7 entity types, and 6 relation types for both the English and Chinese parts of ACE05. We follow \citet{you2022eventgraph} in employing the ACE-E$^{++}$ version of this data, which uses the full text span of entity mentions instead of only the head, as described in section \ref{sec:task-data} above.

For Rich ERE, we keep 18 out of 38 event types defined in the Rich ERE event ontology \footnote{The Rich ERE event ontology defines 38 event types, but for Chinese and Spanish data, only 18 event types are annotated. For consistency, we also use the same 18 event types for the English part.}, 18 out of 21 argument roles \footnote{3 argument roles for the reduced event types are thus excluded.}, 15 entity types, and 6 relation types for English, Chinese, and Spanish. Given no existing data splits, we randomly sample similar proportions of documents for train, development, and testing as the split proportions in ACE05.

\subsection{Evaluation metrics}

Following previous work \citep{lin-etal-2020-joint, nguyen-etal-2021-cross}, precision (P), recall (R), F1 scores are reported for the following information elements.

\begin{itemize}[noitemsep]
    \item \textbf{Entity} An entity mention is correctly extracted if its offsets and entity type match a reference entity.
    \item \textbf{Relation} A relation is correctly extracted if its relation type, and offsets of both entity mentions match those of reference entities.
    \item \textbf{Event trigger} An event trigger is correctly identified (Trg-I) if its offsets match a reference trigger, and correctly classified (Trg-C) if its event type also matches a reference trigger.
    \item \textbf{Event argument} The evaluation of an argument is conditioned on correct event type prediction; if a predicted argument plays a role in an event that does not match any reference event types, the argument is automatically considered a wrong prediction. An argument is correctly identified (Arg-I) if its offsets match a reference argument, and correctly classified (Arg-C) if its argument role also matches the reference argument.
\end{itemize}

\subsection{Implementation detail}

We adopt multi-lingual training for each dataset for the reported results. Results of monolingual models are listed in Appendix \ref{append:monolingual}. Detailed hyper-parameter settings and runtimes are included in Appendix \ref{append:params}.

\subsection{System comparison}

We compare our JSEEGraph to the following systems: 1) ONEIE \citep{lin-etal-2020-joint}; 2) GraphIE \citep{nguyen-etal-2022-joint}; 3) FourIE \citep{nguyen-etal-2021-cross}; 4) JMCEE \citep{xu-etal-2020-novel}; 5) EventGraph \citep{you2022eventgraph} on the ACE05 dataset. For Rich ERE there is little previous work to compare to; the only previously reported results \citep{li2022dual} for EE only solve the task of argument extraction, using gold entity and trigger information, hence their work is not included in our system comparison.

\section{Results and discussion}
\label{sec:results}
We here present the results for our JSEEGraph model for the EE task, as well as its performance for the additional IE components: entities and relations, evaluated as described above. We further zoom in on the nested structures identified in Section \ref{sec:task-data} and assess the performance of our system on these rich structures which have largely been overlooked in previous work on event extraction. We go on to assess the influence of inter-related IE components in an ablation study. Finally we provide an error analysis of our model's predictions.

{\footnotesize
\begin{table}[t!]
\resizebox{\columnwidth}{!}{%
\begin{tabular}{@{}l@{\hspace{2em}}cccc@{\hspace{2em}}cc@{} }
\toprule 
\textbf{Model} & \textbf{Trg-I} & \textbf{Trg-C} & \textbf{Arg-I} & \textbf{Arg-C} & \textbf{Entity} & \textbf{Relation} \\ \midrule

\multicolumn{7}{@{}c@{}}{\raisebox{0.8ex}{\small\textbf{Dataset: ACE05-E$^{+}$ English}  } } \\

EventGraph & \crulefill & 70.0 & \crulefill & 65.4 & \crulefill & \crulefill \\
GraphIE & \crulefill & \textbf{74.8} & \crulefill  & 59.9 & 91.0 & \textbf{65.4} \\ 
ONEIE & 75.6 & 72.8 & 57.3 & 54.8 & 89.6 & 58.6 \\
FourIE & \textbf{76.7} & 73.3 & 59.5 & 57.5 & \textbf{91.1} & 63.6 \\
\textbf{JSEEGraph} & 74.2 & 71.3 & \textbf{70.7} & \textbf{68.4} & 90.7 & 62.6 \\ 
\textbf{JSEEGraph {\footnotesize w/o ent\&rel}} & 74.8 & 71.7 & 67.5 & 64.6 & \crulefill & \crulefill \\ \midrule

\multicolumn{7}{@{}c@{}}{\raisebox{0.8ex}{\small\textbf{Dataset: ACE05-E$^{+}$ Chinese}}} \\
JMCEE &  \textbf{82.3} & \textbf{74.0}& 53.7 & 50 & \crulefill & \crulefill \\
ONEIE &  \crulefill & 67.7 &  \crulefill & 53.2 & \textbf{89.9} & 62.9 \\
FourIE & \crulefill & 70.3 & \crulefill & 56.1 & 89.1 & \textbf{65.9} \\
\textbf{JSEEGraph} & 71.9 & 69.6 & \textbf{74.3} & \textbf{70.1} & 87.4 & 63.3 \\ 
\textbf{JSEEGraph {\footnotesize w/o ent\&rel}} & 70.5 & 67.8 & 69.2 & 65.5 & \crulefill & \crulefill \\ \midrule

\multicolumn{7}{@{}c@{}}{\raisebox{0.8ex}{\small\textbf{Dataset: ACE05-E$^{++}$ English}}} \\

EventGraph & \crulefill & \textbf{74.0} & \crulefill & 58.6 & \crulefill & \crulefill \\
\textbf{JSEEGraph} & 73.5 & 70.0  & \textbf{62.3} &\textbf{59.6}  &  \textbf{85.6} & \textbf{56.6} \\ 
\textbf{JSEEGraph {\footnotesize w/o ent\&rel}} & \textbf{75.0} & 71.3 & 60.3 & 57.7 & \crulefill & \crulefill \\ \midrule

\multicolumn{7}{@{}c@{}}{\raisebox{0.8ex}{\small\textbf{Dataset: ACE05-E$^{++}$ Chinese}}} \\

\textbf{JSEEGraph} & \textbf{69.9} & \textbf{67.8} & \textbf{71.1} & \textbf{66.9} & \textbf{85.2} & \textbf{58.4} \\
\textbf{JSEEGraph {\footnotesize w/o ent\&rel}} & 69.5 & 67.4 & 66.5 & 63.3  & \crulefill & \crulefill \\ \bottomrule

\end{tabular}%
}
\caption{\label{tab:results-ace} \footnotesize Experimental results on ACE05 (F1-score, \%). We bold the highest score of each sub-task.} 
\end{table}}

{\footnotesize
\begin{table}[t!]
\footnotesize
\resizebox{\columnwidth}{!}{%
\begin{tabular}{@{}ll@{\hspace{2em}}cccc@{\hspace{2em}}cc@{} }
\toprule 
\textbf{Model} & \textbf{Lang} & \textbf{Trg-I} & \textbf{Trg-C} & \textbf{Arg-I} & \textbf{Arg-C} & \textbf{Entity} & \textbf{Relation} \\ \midrule

\multicolumn{8}{@{}c@{}}{\raisebox{0.8ex}{\small\textbf{Dataset: Rich ERE-E$^{+}$}  } } \\

\multirow{3}{*}{\textbf{JSEEGraph}} & en & 68.6 & 62.3 & 59.6 & 56.2 & 80.3 & 53.7 \\ 
& zh & 62.7 & 59.0 & 53.1 & 50.1 & 78.1 & 53.2 \\ 
& es & 59.1 & 51.9 & 59.9 & 54.0 & 74.1 & 51.8 \\ \midrule

\multirow{3}{*}{\textbf{JSEEGraph {\footnotesize w/o ent\&rel}}} & en & 67.7 &  62.9 & 57.9 & 54.7 & \crulefill & \crulefill \\
& zh & 63.7 & 60.0 & 50.7 & 48.2 & \crulefill & \crulefill \\ 
& es & 62.3 & 54.3 & 57.3 & 52.5 & \crulefill & \crulefill \\ \midrule

\multicolumn{8}{@{}c@{}}{\raisebox{0.8ex}{\small\textbf{Dataset: Rich ERE-E$^{++}$}  } } \\

\multirow{3}{*}{\textbf{JSEEGraph}} & en & 67.3 & 62.7 & 55.6 & 52.8 & 77.9 & 46.1 \\ 
& zh & 65.2 & 61.7 & 51.0 & 48.7 & 77.5 & 54.3 \\ 
& es & 59.7 & 54.1 & 59.1 & 55.4 & 70.2 & 49.4 \\ \midrule

\multirow{3}{*}{\textbf{JSEEGraph {\footnotesize w/o ent\&rel}}} & en & 66.4 & 61.9 & 52.9 & 50.7 & \crulefill & \crulefill \\ 
& zh & 63.2 & 58.7 & 49.2 & 47.2 & \crulefill & \crulefill \\
& es & 57.2 & 48.9 & 50.8 & 46.4 & \crulefill & \crulefill \\ \bottomrule

\end{tabular}%
}
\caption{\label{tab:results-ere} \footnotesize Experimental results on Rich ERE (F1-score).}
\end{table}}

\subsection{Overall performance}

As shown in Table \ref{tab:results-ace}, on ACE-E$^+$, our overall results align with other systems. Our JSEEGraph results are especially strong for event argument extraction, with an improvement of around 10 percentage points from the best results of the previous best performing systems in our comparison.

On the newly introduced ACE-E$^{++}$, despite having more complex structures, with a higher degree of nested  structures, the results of JSEEGraph on trigger extraction remain stable. We further note that our results on argument, entity, and relation extraction suffers some loss from highly nested entities, which is not surprising. 

From Table \ref{tab:results-ere}, we find that the scores on Rich ERE are consistently lower compared to those of ACE05. The double-tagging of event triggers described in Section \ref{sec:task-data} clearly pose a certain level of difficulty for the model to disambiguate events with a shared trigger. Argument and entity extraction also suffers from more fined-grained entity types.

\subsection{Nesting}

In order to directly evaluate our model's performance on nested instances, we split each test set into nested and non-nested parts and report the corresponding scores, as shown in Table \ref{tab:results-nesting}\footnote{ACE05-E$^+$ is not included as it lacks sufficient nested instances.}. 

We observe that JSEEGraph is quite robust in tackling nested instances across different IE tasks and languages. On ACE05-E$^{++}$, more than half of the test data are nested for both English and Chinese, and the results on the nested parts are lower, however consistently comparable with the non-nested parts of the datasets. On Rich ERE-E$^{+}$, nested instances make up only a small part of the test data, but the results are still comparable to the non-nested part. On Rich ERE-E$^{++}$, about one third of the test data are nested, results of the nested parts are in fact consistently better for trigger, entity, and relation extraction, but inferior for argument extraction.

To conclude, JSEEGraph does not suffer considerable performance loss from nesting among different IE elements, and in many cases actually gains in performance from more complex structures, notably for trigger, entity, and relation extraction. It is clear that the system can make use of inter-relations between the different IE elements of the information graph in order to resolve these structures.

\subsection{Ablation study}
In order to gauge the effect of the joint modeling of entities, events, and relations, we perform an ablation study where we remove the entity and relation information from our information graph, hence only performing the task of event extraction directly from text. In the reduced information graph, node labels for entity types are removed, and relation edges between entities are also removed. We find that event extraction clearly benefits from entity and relation extraction, especially for event argument extraction. As shown in Table \ref{tab:results-ace} and Table \ref{tab:results-ere}, when we train our model only for event extraction, the performance on argument extraction drops consistently across different datasets and languages, but the performance on trigger extraction remains quite stable.

{\footnotesize
\begin{table}[t!]
\footnotesize
\resizebox{\columnwidth}{!}{%
\begin{tabular}{@{}lll@{\hspace{2em}}cccc@{\hspace{2em}}cc@{} }
\toprule 
\textbf{Lang} & \textbf{Nested} & \textbf{\#sents} & \textbf{Trg-I} & \textbf{Trg-C} & \textbf{Arg-I} & \textbf{Arg-C} & \textbf{Entity} & \textbf{Relation} \\ \midrule

\multicolumn{9}{@{}c@{}}{\raisebox{0.8ex}{\small\textbf{Dataset: ACE05-E$^{++}$ }  } } \\

\multirow{2}{*}{\textbf{en}} & \ding{51} & 418 & 72.1 & 68.5 & 59.2 & 57.0 & 85.1 & 57.0 \\
& \ding{55} & 359 & 77.0 & 74.0 & 73.2 & 69.0 & 87.4 & 47.5 \\  

\multirow{2}{*}{\textbf{zh}} & \ding{51} & 277 & 72.2 & 69.7 & 68.9 & 65.5 & 85.4 & 60.8\\
& \ding{55} & 244 & 57.6 & 57.6 & 87.9 & 77.3 & 84.5 &  33.6 \\ \midrule

\multicolumn{9}{@{}c@{}}{\raisebox{0.8ex}{\small\textbf{Dataset: Rich ERE-E$^{+}$ }  } } \\

\multirow{2}{*}{\textbf{en}} & \ding{51} & 93 & 81.3 & 71.6 & 54.8 & 51.4 & 81.3 & 49.8 \\
& \ding{55} & 652 & 61.4 & 56.9 & 64.0 & 60.5 & 79.8 & 56.3 \\ 

\multirow{2}{*}{\textbf{zh}} & \ding{51} & 101 & 72.0 & 66.6 & 47.5 & 45.1 & 79.7 & 56.0  \\
& \ding{55} & 382 & 54.2 & 52.2 & 59.5 & 55.9 & 77.1 & 49.9 \\ 

\multirow{2}{*}{\textbf{es}} & \ding{51} & 51 & 78.1 & 64.7 & 55.5 & 52.3 & 78.4 & 51.8 \\
& \ding{55} & 547 & 49.9 & 45.5 & 63.8 & 55.6 & 73.1 & 51.8 \\ \midrule

\multicolumn{9}{@{}c@{}}{\raisebox{0.8ex}{\small\textbf{Dataset: Rich ERE-E$^{++}$ }  } } \\

\multirow{2}{*}{\textbf{en}} & \ding{51} & 251 & 75.4 & 69.2 & 53.0 & 50.5 & 81.0 & 45.7 \\
& \ding{55} & 494 & 46.0 & 45.3 & 75.0 & 70.6  & 71.0 & 49.4 \\ 

\multirow{2}{*}{\textbf{zh}} & \ding{51} & 197 & 70.4 & 67.0 & 49.0 & 46.8 & 80.4 & 57.2\\
& \ding{55} & 286 & 45.9 & 41.8 & 63.9 & 61.1 & 69.7 & 23.3 \\ 

\multirow{2}{*}{\textbf{es}} & \ding{51} & 163 & 66.0 & 59.3 & 57.2 & 53.7 & 75.2 & 53.5 \\
& \ding{55} & 435 & 47.0 & 43.0 & 65.3 & 61.7 & 61.4 & 30.0 \\ \bottomrule
\end{tabular}%
}
\caption{\label{tab:results-nesting} \footnotesize Experimental results on test data with nesting as compared to without nesting (F1-score, \%).} 
\end{table}}

\subsection{Error analysis}

The experimental results show that JSEEGraph has an advantage when it comes to the task of  argument extraction. In a manual error analysis we therefore focus on the errors of event trigger extraction. After a manual inspection of our model's predictions on the test data, we find that the errors fall into the following main categories.

\paragraph{\textbf{Over-predict non-event sentences.}} Our system tends to be more greedy in extracting event mentions, and wrongly classifies some tokens as event triggers even though the sentence does not contain event annotation. For instance, the sentence \textit{``Anne-Marie will get the couple's 19-room home in New York state''} (from ACE05) does not have annotated events, but our system extracts \textit{``get''} as trigger for a \texttt{Transfer-Ownership} event; in this case, however, one could argue that the \texttt{Transfer-Ownership} should be annotated. 

\paragraph{\textbf{Under-predict multi-event sentences}} When a sentence contains multiple event mentions, JSEEGraph sometimes fails to extract all of the event triggers. For example, this sentence \textit{``Kelly , the US assistant secretary for East Asia and Pacific Affairs , arrived in Seoul from Beijing Friday to brief Yoon, the foreign minister''} from ACE05 contains a \texttt{Transport} event triggered by \textit{``arrived''} and a \texttt{Meet} event triggered by \textit{``brief''}, but our system fails to extract the trigger for the \texttt{Meet} event; in this example, it requires a certain level of knowledge to be able to identify \textit{``brief''} as an event trigger, which is beyond the capacity of our model.

\paragraph{\textbf{Wrong event types}} In some cases, even though our model successfully identifies an event trigger, it assigns a wrong event type. Some event types can easily be confused with each other. In this sentence from Rich ERE, \textit{``The University of Arkansas campus was buzzing Friday after a student hurt himself when a gun went off in his backpack in the KUAF building''}, an \texttt{Injure} event is evoked by \textit{``hurt''}, but our model assigns an event type of \texttt{Attack}. Clearly, \texttt{Injure} and \texttt{Attack} events are one typical case of event types that can be easily confused.

\paragraph{\textbf{Context beyond sentence}} This error applies specifically to Rich ERE: even though the annotation of events is on a sentence level, annotators were instructed to take into account the context of the whole article. Our model fails completely when a trigger requires context beyond the sentence. For instance, this sentence \textit{``If Mickey can do it , so can we!''} is taken from an article describing an on-going demonstration in Disney Land, and \textit{``it''} is the trigger for a \texttt{demonstrate} event; without the context, our model fails to identify the trigger. These are cases which would require information about event coreference.

\section{Conclusion}
\label{sec:conclusion}

In this paper, we have proposed JSEEG, a graph-based approach for joint structured event extraction, alongside entity, and relation extraction. We experiment on two benchmark datasets ACE05 and Rich ERE, covering the three languages English, Chinese, and Spanish. We find that our proposed JSEEGraph is robust in solving nested event structures, and is especially strong in event argument extraction. We further demonstrate that it is beneficial to jointly perform EE with other IE tasks, and event argument extraction especially gains from entity extraction.

\section*{Limitations}
\label{sec:limitations}

Our work has two main limitations. Firstly, we do not compare our system to previous works on the Rich ERE dataset. This is mainly due to the fact that most work  use the light ERE \citep{song-etal-2015-light} dataset. We were unfortunately not able to got access to this version of the data\footnote{Here we refer to the datasets with LDC codes: \textit{LDC2015E29}, \textit{LDC2015E68}, and \textit{LDC2015E78} for English ERE, and \textit{LDC2015E107} for the Spanish ERE.}, which is why no experiments were carried out on it. 

Secondly, we only experiment with one language model, the multilingual model XLM-R. As our model is language agnostic, and we aimed to test its performance on datasets in different languages, the choice of a multilingual model was obvious. XLM-R has been chosen based on its good performance in other tasks, and to make our work comparable to previous work \cite{you2022eventgraph}.
However, another approach would be to test our model with a selection of language-specific language models. 

\section*{Acknowledgements}

This work was supported by industry partners and the Research Council of Norway with funding to \textit{MediaFutures: Research Centre for Responsible Media Technology and Innovation}, through the centers for Research-based Innovation scheme, project number 309339.


\bibliography{anthology,custom}
\bibliographystyle{acl_natbib}
\clearpage
\newpage
\appendix

\section{Training detail}
\label{append:params}

We use the large version of XLM-R available on HuggingFace \texttt{transformers}\footnote{\url{https://huggingface.co/docs/transformers/index}} for obtaining contextual embeddings of the input sequence. We use the same hyperparameter configuration for all our models, as shown in Table \ref{tab:hyperparams}, and weights are optimized with AdamW \citep{loshchilov2017decoupled} following a warmed-up cosine learning rate schedule.

\begin{table}[!h]
\resizebox{\columnwidth}{!}{%
\begin{tabular}{@{}lr@{}}
\toprule
\textbf{Hyperparameter} & \textbf{JSEEGraph} \\ \midrule
batch\_size & 16 \\                    
beta\_2 & 0.98 \\                    
decoder\_learning\_rate & 1.0e-4 \\      
decoder\_weight\_decay & 1.2e-6 \\        
dropout\_transformer & 0.25 \\         
dropout\_transformer\_attention & 0.1 \\
encoder & \textit{"xlm-roberta-large"} \\      
encoder\_learning\_rate & 4.0e-6 \\   
encoder\_weight\_decay & 0.1 \\      
epochs & 110 \\            
hidden\_size\_anchor & 256 \\          
hidden\_size\_edge\_label & 256 \\      
hidden\_size\_edge\_presence & 256 \\   
n\_transformer\_layers & 3 \\                      
query\_length & 2 \\        
warmup\_steps & 1\,000 \\ \bottomrule
\end{tabular}%
}
\caption{Hyperparameter setting for our system, and we use the same configuration for all models.}
\label{tab:hyperparams}
\end{table}

The training was done on a single node of Nvidia RTX3090 GPU. The runtimes and sizes (including the pretrained XLM-R) of the multilingual models for each dataset are listed in Table \ref{tab:runtime}, 

\begin{table}[!h]
\resizebox{\columnwidth}{!}{%
\begin{tabular}{@{}l@{\hspace{3em}}r@{\hspace{2em}}r@{}}
\toprule
\textbf{Dataset} & \textbf{Runtime} & \textbf{Model size} \\ \midrule                  
ACE05-E${^+}$ & 27:52 h & 343.8 M \\  
ACE05-E${^{++}}$ & 27:25 h & 343.8 M \\
Rich ERE-E${^{+}}$ & 33:13 h & 344.6 M \\
Rich ERE-E${^{++}}$ & 32:16 h & 344.6 M \\ \bottomrule
\end{tabular}%
}
\caption{The training times and model sizes (number of trainable weights) of all our experiments.}
\label{tab:runtime}
\end{table}

\newpage

\section{Monolingual training results}
\label{append:monolingual}

Apart from multilingual training, we also train two monolingual models for each language, one for joint event extraction with entity and relation and the for event extraction only. Results of monolingual models are summerized in Table \ref{tab:results-mono}.

\begin{table}[t!]
\resizebox{\columnwidth}{!}{%
\begin{tabular}{@{}l@{\hspace{2em}}cccc@{\hspace{2em}}cc@{} }
\toprule 
\textbf{Lang} & \textbf{Trg-I} & \textbf{Trg-C} & \textbf{Arg-I} & \textbf{Arg-C} & \textbf{Entity} & \textbf{Relation} \\ \midrule

\multicolumn{7}{@{}c@{}}{\raisebox{0.8ex}{\small\textbf{Dataset: ACE05-E$^{+}$}  } } \\

\multirow{2}{*}{\textbf{en}} & 73.1 & 70.0 & 68.5 & 65.4 & 90.4 & 61.4 \\
 & 73.2 & 69.8 & 66.7 & 64.2 & \crulefill & \crulefill \\ 

 \multirow{2}{*}{\textbf{zh}} & 69.2 & 67.0 & 71.4 & 67.8 & 85.6 & 60.2 \\
 & 64.8 & 62.6 & 62.5 & 59.3 & \crulefill & \crulefill \\ \midrule

\multicolumn{7}{@{}c@{}}{\raisebox{0.8ex}{\small\textbf{Dataset: ACE05-E$^{++}$}}} \\

\multirow{2}{*}{\textbf{en}} & 73.8 & 70.3 & 63.7 & 60.6 & 85.3 & 55.4 \\
 & 72.7 & 69.9 & 58.9 & 56.3 & \crulefill & \crulefill \\ 

 \multirow{2}{*}{\textbf{zh}} & 66.7 & 64.5 & 66.0 & 63.1 & 82.1 & 53.7 \\
 & 66.0 & 64.3 & 62.2 & 58.4 & \crulefill & \crulefill \\ \midrule

\multicolumn{7}{@{}c@{}}{\raisebox{0.8ex}{\small\textbf{Dataset: Rich ERE-E$^{+}$}}} \\

\multirow{2}{*}{\textbf{en}} & 65.3 & 60.5 & 59.8 & 56.1 & 80.6 & 53.6 \\
 & 68.7 & 62.4 & 56.0 & 52.8 & \crulefill & \crulefill \\ 

 \multirow{2}{*}{\textbf{zh}} & 62.3 & 57.7 & 53.9 & 50.2 & 78.3 & 54.5 \\
 & 62.4 & 59.0 & 48.2 & 46.3 & \crulefill & \crulefill \\ 

 \multirow{2}{*}{\textbf{es}} & 54.2 & 47.9 & 52.5 & 46.7 & 72.9 & 44.7 \\
 & 56.7 & 49.7 & 51.3 & 47.3 & \crulefill & \crulefill \\ \midrule

\multicolumn{7}{@{}c@{}}{\raisebox{0.8ex}{\small\textbf{Dataset: Rich ERE-E$^{++}$}}} \\

\multirow{2}{*}{\textbf{en}} & 66.9 & 60.4 & 54.6 & 52.1 & 76.3 & 42.1 \\
 & 66.2 & 59.2 & 49.5 & 46.8 & \crulefill & \crulefill \\ 

 \multirow{2}{*}{\textbf{zh}} & 63.6 & 60.2 & 47.1 & 44.8 & 76.2 & 51.5 \\
 & 60.5 & 57.2 & 41.5 & 38.8 & \crulefill & \crulefill \\ 

 \multirow{2}{*}{\textbf{es}} & 54.4 & 48.9 & 47.2 & 43.2 & 68.2 & 43.0 \\
 & 54.5 & 48.4 & 35.7 & 32.1 & \crulefill & \crulefill \\ \bottomrule

\end{tabular}%
}
\caption{\label{tab:results-mono} Experimental results of monolingual models (F1-score, \%)}
\end{table}

\end{document}